\documentclass[conference]{IEEEtran}
\IEEEoverridecommandlockouts
\usepackage{cite}
\usepackage{amsmath,amssymb,amsfonts}
\usepackage{algorithmic}
\usepackage{graphicx}
\usepackage{textcomp}
\usepackage{xcolor}
\usepackage{multirow}
\def\BibTeX{{\rm B\kern-.05em{\sc i\kern-.025em b}\kern-.08em
    T\kern-.1667em\lower.7ex\hbox{E}\kern-.125emX}}
\begin{document}

\title{Misspelling Correction with Pre-trained Contextual Language Model\\

\thanks{}
}

\makeatletter 
\newcommand{\linebreakand}{
\end{@IEEEauthorhalign}
\hfill\mbox{}\par
\mbox{}\hfill\begin{@IEEEauthorhalign}
}
\makeatother

\author{\IEEEauthorblockN{Yifei Hu}
\IEEEauthorblockA{\textit{Computer and Information Technology} \\
\textit{Purdue University}\\
hu381@purdue.edu}
\and
\IEEEauthorblockN{Xiaonan Jing}
\IEEEauthorblockA{\textit{Computer and Information Technology} \\
\textit{Purdue University}\\
jing@purdue.edu}
\and
\IEEEauthorblockN{Youlim Ko}
\IEEEauthorblockA{\textit{Computer and Information Technology} \\
\textit{Purdue University}\\
ko102@purdue.edu}
\linebreakand
\IEEEauthorblockN{Julia Taylor Rayz}
\IEEEauthorblockA{\textit{Computer and Information Technology} \\
\textit{Purdue University}\\
jtaylor1@purdue.edu}
}

\maketitle

\begin{abstract}
Spelling irregularities, known now as spelling mistakes, have been found for several centuries. As humans, we are able to understand most of the misspelled words based on their location in the sentence, perceived pronunciation, and context. Unlike humans, computer systems do not possess the convenient auto complete functionality of which human brains are capable. While many programs provide spelling correction functionality, many systems do not take context into account. Moreover, Artificial Intelligence systems function in the way they are trained on. With many current Natural Language Processing (NLP) systems trained on grammatically correct text data, many are vulnerable against adversarial examples, yet correctly spelled text processing is crucial for learning. In this paper, we investigate how spelling errors can be corrected in context, with a pre-trained language model BERT. We present two experiments, based on BERT and the edit distance algorithm, for ranking and selecting candidate corrections. The results of our experiments demonstrated that when combined properly, contextual word embeddings of BERT and edit distance are capable of effectively correcting spelling errors. 

\end{abstract}

\begin{IEEEkeywords}
Spelling Correction, BERT, Contextual Language Model, Natural Language Processing
\end{IEEEkeywords}

\section{Introduction}
Language is a unique gift for humankind. It conveys information using rules with a certain amount of informality. Human can quickly adapt to the informality and develop new rules to make sense from the informality. However, understanding the informality in human language has been a challenging task for computer algorithms. Nowadays, many state-of-the-art Natural Language Processing (NLP) tools and architectures are still vulnerable against all sorts of informality. 

A simple spelling error can greatly impact the output of a sophisticated natural language processing tool for tasks that enjoy very small error rates. For example,
Named Entity Recognition (NER) is one of the common NLP tasks, which automatically identifies named entities from a given text input. The following example is taken from  AllenNLP \cite{b5}  ELMO-based \cite{b6} online NER demo:

\begin{center}
``\emph{this shirt was bought \textbf{at} Grandpa Joe's in LA.}'' 
\end{center}

In this sentence, two named entities, \emph{Grandpa Joe's} (as an organization) and \emph{LA} (as a city), are identified. However, if the input is slightly modified by introducing a misspelled word  ---  ``\emph{at}'' changed to ``\emph{ta}'' (which is a common misspelling for ``\emph{at}'' ) --- an utterance``\emph{Grandpa Joe's}'', which was previously identified as an organization, will not no longer be identified as a named entity. Instead, ``\emph{Joe}'', by itself, will be identified as a person.

Similarly, spelling correction affects another common NLP task, the semantic role labeling (SLR). Consider the following example: 
\begin{center}
``\emph{the keys, which \textbf{were} needed to access the building, were locked in the car}'' 
\end{center}

SLR model based on BERT (Bidirectional Encoder Representations from Transformers) \cite{b1} identifies ``the keys" as the only argument in this sentence.   However, if the first word ``\emph{were}'' is misspelled as ``\emph{weer}'', the misspelled ``\emph{weer}'' would be captured as a second argument.

For human, the spelling errors above do not impact how the sentences are interpreted. However, both the ELMO-based model and the BERT-based model changed their outputs as a result of misspellings present in the sentences. Similar to the above examples, many current NLP models are not robust enough when processing flawed input data. However, corpora collected through human generated texts often contain errors such as spelling errors, incorrect usage of words, and grammatical mistakes. These mistakes can easily introduce noise to the corpora and affect NLP parsers used for subsequent tasks. In many cases, minimizing noise in the corpora by correcting unintentional human generated mistakes serves an important procedure in performance enhancement for machine learning models. 

This paper focuses on contextual correction of misspelled words --- one of the most common type of errors in natural language texts. Spelling errors can be roughly divided into two types, non-word error and real-word error \cite{b7}. The former type results in a misspelled mistake which is not an actual English word (i.e. "weer" in the previous example) and the latter results in a mistake that is accidentally an English word (i.e. misspell "access" as "assess"), but does not work well in context. While the detection of non-word error can be employed through a look-up in a large English vocabulary dictionary, the detection of real-word errors is more challenging. Real-word error detection often requires training of expensive neural network models with a large gold standard corpora in order to locate the desired error. Thus, training a robust error encoder which detects both types of errors can be difficult. On the other hand, both the correction of non-word and real-word errors can share a universal framework as both errors are a result of unintentional representation of the intended word. For instance, both "assess" (real-word error) and "acsese" (non-word error) are two edit distance away from the intended word "access" in the previous example. Fundamentally, the efforts spent in correcting both misspelled words to the intended word "access" are identical. With this intuition, we introduce a simple approach to correcting spelling errors in this paper which utilizes the edit distance mechanism and the recently popular neural network model BERT. 

BERT uses transformer \cite{b16} architecture to predict masked word given the context. The attention mechanism in the transformers allows the model to compute an attention score for each context word and subsequently use the attention score to predict the masked word. We are interested in examining BERT's ability to predict correct words given the masked spelling errors.

\section{Related Works}

\subsection{Traditional Spelling Error Correction}

Traditional spelling error correction approaches include noisy channel models and n-gram models. Church and Gale \cite{b8} presented a context-based system with a noisy channel model, employing a simple word bi-gram model and Good-Turing frequency estimation. A similar model from IBM Watson Research using tri-gram was proposed in \cite{b9}. Carlson and Fette \cite{b10} illustrated the effectiveness of n-gram models in spelling error corrections. They compared the n-gram models with the GNU Aspell \cite{b11} model on Google n-gram dataset. On non-word error corrections, n-gram model showed a 36\%-49\% improvements on top correct word ranking, with highest accuracy achieved at 92.4\% for detecting insertion error with 5-gram. A recent study on real-time spelling error correction proposed a Symmetric Delete Algorithm (SDA) with n-grams approach \cite{b12}. A weighted sum of unigrams, bigrams, and trigrams was used as the metric for ranking SDA generated candidate words. The model was tested on 24 languages with spelling errors generated by manipulating edit distance and bi-gram probability, with a 97.1\% accuracy on top English language errors. However, the model showed only 68\% top correction accuracy when tested on public dataset such as Wikipedia, although it outperforms popular industrial models Aspell and Hunspell \cite{b13}.

\subsection{Neural Network Based Spelling Error Correction}
Sequence-based neural network models have also been applied in spelling error correction tasks. Li et al. trained a nested RNN (Reccurent Neural Network) model with a large-scale pseudo dataset generated from phonetic similarity \cite{b14}. The model outperforms other state-of-the-art sequential models, scRNN \cite{b23} and LSTM-Char-CNN \cite{b24} -- which combined long short-term memory (LSTM) and character-level convolutional neural networks (Char-CNN) -- by 5\% on precision and 2\% on recall. Ge et al. trained a neural sequence-to-sequence model in grammatical error correction (GEC) \cite{b15}. Through fluency boost mechanism, which allows multi-round sequence-to-sequence corrections, multiple grammatical errors in one sentence can be corrected in one setting. However, such training often requires large amount of data to enhance the performance of the error encoder. Oftentimes, spelling error dataset with gold standard labels are small on size and generated errors can be drastically different from how human writers make mistakes. In a recent attempt, Shaptala and Didenko \cite{b17} proposed an alternative GEC approach using pre-trained BERT with fully connected network to detect and correct errors. However, the error type resolution given by the model does not consider the dependency between different errors and error types, and the decoder tends to mistakenly remove end of sentence tokens. The result of \cite{b17} again demonstrated the difficulties in detecting errors without type independence and the limitation of neural network decoders. 




\section{Dataset}
The dataset used in the paper is a subset from the CLC (Cambridge Learner Corpus) FCE (First Certificate in English) Dataset \cite{b18}, which consists of 5,124 sentences written by English as second language learners, from speakers of 16 native languages. CLC FCE dataset contains 75 types of annotated errors and corresponding corrections in the original essays. We extracted a subset which contains only spelling errors, annotated as "S", "SA", "SX" to represent "spelling", "American spelling", and "spelling confusion" errors respectively. A total of 2,075 sentences was extracted from the original dataset. Each extracted sentence was limited to contain only one misspelled word in order to provide BERT with the correct contextual information. TABLE I shows the distribution of the dataset. We used the Brown corpus (49815 vocabulary) from NLTK (Natural Language Toolkit) \cite{b21} to determine whether the error was a real-word.

\begin{table} [h]
\caption{dataset statistics}
\label{T1}
\begin{center}
\begin{tabular}{ | c || c || c | } 
 \hline
 \multirow{4}{4em}{\textbf{sentence statistics}} & average sentence length & 20.68 \\ 
 & std of sentence length & 11.00 \\ 
 & maximum length & 1  \\ 
 & minimum length & 99 \\ \hline
 \multirow{2}{4em}{\textbf{error types}} & number of real-word errors & 225 \\
 & number of non-word errors & 1850 \\ \hline 
\end{tabular}
\end{center}
\end{table}

\section{Proposed Method}
We propose two methods which combine the per-trained BERT language model and the edit distance algorithm. The pre-trained BERT model is capable of predicting a masked word and provide a list of candidate words that make sense in a given context.  The edit distance algorithm can be used to find the words that are similar to the misspelled word. The detail of this mechanism will be explained in the following section. 

\subsection{Treating Misspelled Words As Masked Words}\label{AA}
BERT provides a functionality of predicting a word in a sentence, marked as needed to be predicted,  based on the context of a sentence. Due to this mechanism, known as masked word prediction, BERT only takes a sentence with one masked token (word) as input at a time and output a list of candidate words with probabilities accordingly from high to low. Here is an example of a typical masked word prediction process:

\begin{enumerate}
\item Raw sentence: ``\emph{How are you today?}''
\item Replace a word with a MASK: ``\emph{How [MASK] you today?}''
\item Predictions (with probability) of the MASK: ``\emph{are (0.88)}'', ``\emph{do (0.04)}'', ``\emph{about (0.03)}'', ``\emph{have (0.003)}''
\end{enumerate}

Based on this mechanism, it is possible to encode the misspelled words as masked words in each sentence and utilize the one to one prediction feature in candidate correction rankings. The number of candidate words is adjusted by a parameter N which can be as large as the vocabulary size. However, one limitation for the BERT model is that if the expected output is not in BERT's vocabulary, BERT will not be able to give the correct predictions that would match a gold standard. In reality, BERT is trained on various corpora with diverse vocabulary, which makes it very unlikely for BERT to encounter an out-of-vocabulary word. In this paper, we uses a pre-trained BERT-Large-Cased model \cite{b20} in candidate correct words generation. 

It should be noted that BERT and similar architectures are not the only mechanisms of providing an unknown word based on context. An ontology-based approach was described by Taylor and Raskin \cite{b19} by using fuzzy membership functions of words based on the context of a sentence. The differences between human and computer processing of information in context was also touched on by Jing et al. \cite{b22}. 

\subsection{Damerau–Levenshtein Distance}\label{AA}
Edit distance \cite{b3}\cite{b4} measures the similarity between two words syntactically. In this paper, edit distance refers to the Damerau–Levenshtein Distance. There are 4 different operations: insertion, deletion, substitution, and transposition. Below are the examples for each type of operation:

\begin{itemize}
\item Insertion: ``\emph{wat}'' $\,\to\,$ ``\emph{what}''
\item Deletion: ``\emph{whaat}'' $\,\to\,$ ``\emph{what}''
\item Substitution: ``\emph{wgat}'' $\,\to\,$ ``\emph{what}''
\item Transposition: ``\emph{waht}'' $\,\to\,$ ``\emph{what}''
\end{itemize}

Edit distance is calculated as the count of a minimum number of operations above needed to covert one word to another.

In this paper, edit distance is used in two different ways, comparison and generation. First, edit distance is used to compare the candidate corrections from the BERT prediction to the misspelled word. Ideally, the corrections should be very similar to the misspelled words in terms of edit distance. The candidate correction with the lowest edit distance is selected as the correct prediction of the misspelled word. It is possible that the lower edit distance is selected for a word that is also slightly lower on the BERT's list in terms of predictions than a word with a higher edit distance. However, since all BERT's predictions are contextual, and people tend to recover similar words faster, we choose to optimize edit distance rather than contextual ranking.  Second, edit distance is used to find all similar words of a target token. Given a corpus, we can extract existing lexical units that are within K edit distance away from the given word and then test it with BERT. For instance, in the Brown corpus, all the words close to "annoying" with an edit distance under 2: enjoying, annoying. These two approaches can be seen as bidirectional: one takes BERT's interpretation of misspelled words and then applies edit distance to it; the other one applies edit distance first, and then checks whether it agrees with BERT's interpretation of what fits well in a sentence. 

\section{Experiments and Results}
\subsection{Experiment 1: Applying BERT before edit distance}\label{AA}
In this experiment, we first apply the pre-trained BERT-Large-Cased \cite{b20} model to predict the masked misspelled words, then employ the edit distance to rank the candidate corrections. The BERT model outputs a list of N candidate words, where we chose N to be from 10 to 500 for this experiment. The edit distance between each candidate word and the misspelled token can be then calculated and ranked in ascending order. The word with the lowest edit distance is chosen as the final prediction. When multiple candidate words have the same lowest edit distance, the word with the highest prediction probability from BERT is chosen as the final prediction.



The result for experiment 1 is shown in Table II. Three metrics are applied in evaluating the accuracy of the corrections:

\begin{itemize}
    \item \textbf{accuracy top@1}: measure how often the label matches with top 1 prediction. 
    \item \textbf{accuracy top@N}: measures how often the label appears in the top N output of BERT.
    \item \textbf{P(top@1$\vert$top@N)}: measures the conditional probability for the edit distance to select top 1 correct prediction given the label appears in top N output of BERT.
\end{itemize}

\begin{table}[h]
\caption{Results for Experiment 1, based on various number of BERT predictions, N}
\label{table_example}
\begin{center}
\begin{tabular}{|c||c||l||c|}
\hline
top-N & \begin{tabular}[c]{@{}c@{}}accuracy\\ top@1\end{tabular} & \begin{tabular}[c]{@{}l@{}}accuracy\\ top@N\end{tabular} & P(top@1$\vert$top@N) \\ \hline
10    & 48.77\%                                                  & 49.64\%                                                  & 98.25\%       \\ \hline
50    & 64.34\%                                                  & 66.80\%                                                  & 96.32\%       \\ \hline
100   & 68.34\%                                                  & 71.71\%                                                  & 95.30\%       \\ \hline
200   & 71.08\%                                                  & 75.52\%                                                  & 94.13\%       \\ \hline
300   & 72.14\%                                                  & 77.20\%                                                  & 93.45\%       \\ \hline
400   & 72.92\%                                                  & 78.17\%                                                  & 93.28\%       \\ \hline
500   & \textbf{73.25\%}                                                  & 78.84\%                                                  & 92.91\%       \\ \hline
\end{tabular}
\end{center}
\end{table}

\begin{figure}[thpb]
  \centering
  \framebox{\parbox{3in}{\includegraphics[scale=0.17]{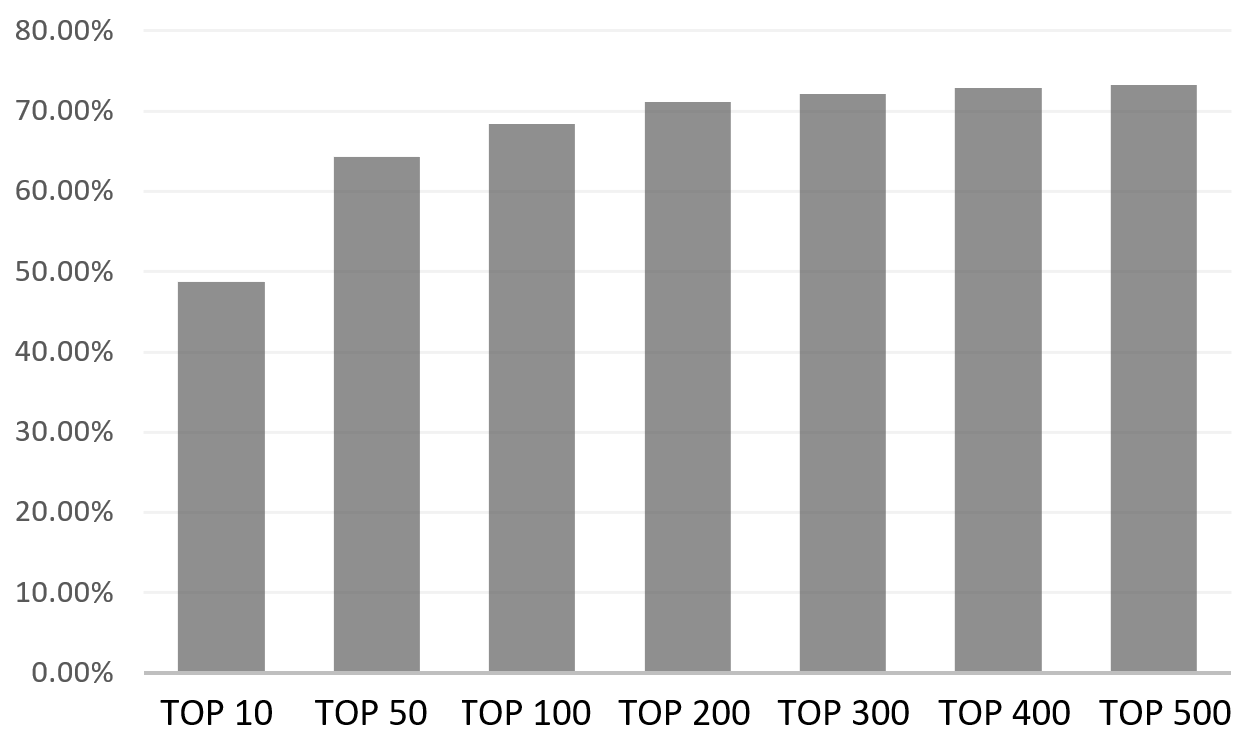}}}
  
  \caption{Final accuracy changes when increasing N}
  \label{figurelabel}
\end{figure}

With N = 500, 73.25\% of the misspellings were corrected. According to Fig. 1, by increasing N, the final accuracy increased accordingly; however, the trend was slowing down. At the same time, as we increased N, the edit distance algorithm started to miss more correct answers from the candidate list. The metric P(top@1$\vert$top@N) showed a clear decreasing trend. Combining the accuracy top@1 and the P(top@1$\vert$top@N), we can expect that the accuracy top@1 may converge.

\subsection{Experiment 2: Applying BERT after edit distance}\label{AA}
As stated earlier, in this experiment we first generate a list of similar words within the K edit distance of the misspelled token, then apply BERT to rank the candidate corrections. BERT is capable of computing a softmax probability for each candidate word within the provided context. Thus, the word with the highest probability is chosen as the final prediction. 

As mentioned in the previous section, edit distance can also be used as a criteria to filter lexical entities from an existing corpus. Without the corpus restriction, many non-word contents are generated through modifying the misspelled token, which leaves BERT's vocabulary as the only criteria to restrict real-word selection. When a particular corpus is applied prior to BERT's selection, it is feasible for common words or domain specific vocabularies to be captured to further improve the accuracy of the corrections. 

Note that in experiment 1, a threshold is not required for edit distance to realize the selection process as the minimal value is always preferred. In this experiment, as the edit distance is applied as the first round selection criteria, a threshold is required since it is impossible to generate candidates with an infinitely large K (one could keep adding letters to a word). Such restriction of K can affect the accuracy of the corrections, since the labels can be out-of-vocabulary words which are filtered out before the subsequent steps. 

In order to compare how edit distance and corpus can jointly affect the candidate selections, we control the example sentences sampled from the dataset to eliminate out-of-vocabulary words. Three subsets of sentences are created based on the following mechanism:
\begin{itemize}
    \item sample the sentence if the label is within K edit distance of the misspelled token. 
    \item sample the sentence if the label is also in the corpus. 
    \item sample the sentence if the label is within K edit distance of the misspelled token and that the word is in the corpus. 
\end{itemize}

In this particular experiment, an edit distance of K $\leq$ 2 and the Brown corpus are used for all sub-tasks. Among the 2075 testing cases, 1934 of them have K $\leq$ 2. If we increased the K to 3 or 4, the coverage barely grows but the cost of computation can grow exponentially. TABLE III shows some examples of words and the number of similar spellings with different edit distance:

\begin{table}[h]
\caption{Number of similar words with different edit distance (ED)}
\label{table_example}
\begin{center}
\begin{tabular}{|c||c||c||c||c|}
\hline
Word        & ED = 1 & ED = 2 & ED = 3 & ED = 4 \\ \hline
study       & 4      & 35     & 427    & 3148   \\ \hline
annoying    & 1      & 2      & 42     & 480    \\ \hline
adventurous & 1      & 3      & 7      & 13     \\ \hline
\end{tabular}
\end{center}
\end{table}

The top@1 accuracy for each restricted subsets, as described below, are compared with a comparison group with no restrictions on either edit distance or corpus. The results are compared between the following groups:
\begin{itemize}
\item \textbf{no restriction}: no sub-sampling on the dataset. 
\item \textbf{ED2 only}: subset containing examples if the label is within 2 edit distance of the misspelled token. 
\item \textbf{corpus only}: subset containing examples if the label is also in the Brown corpus.
\item \textbf{ED2 + corpus}: subset containing examples if the label is within 2 edit distance of the misspelled token and that the word is in the Brown corpus. 
\end{itemize}

TABLE IV shows the result for each group.

\begin{table}[h]
\caption{Experiment 2 result}
\label{table_example}
\begin{center}
\begin{tabular}{|c||c||c||c|}
\hline
                     & Total Cases                  & Correct Cases          & Accuracy \\ \hline
No Restriction       & {\color[HTML]{000000} 2075} & {\color[HTML]{000000} 1491} & 71.86\%  \\ \hline
ED2  only & 1934                        & 1491                        & 77.09\%  \\ \hline
corpus  only    & 1885                        & 1491                        & 79.10\%  \\ \hline
ED2  +  corpus  & 1756                        & 1491                        & 84.91\%  \\ \hline
\end{tabular}
\end{center}
\end{table}

\section{Discussion}

\subsection{BERT can be used for misspelling correction}

In this experiment, the BERT model was not specially trained or fine-tuned. It was not designed for misspelling correction task but still provided acceptable result in both experiments. After applying the edit distance algorithm with the BERT model, 73.25\% of the misspellings were corrected successfully in experiment 1. If we exclude the out-of-vocabulary words and the errors with large edit distance in experiment 2, the accuracy can reach 84.91\%. The experiments showed using BERT to do misspelling correction was possible if we treated the misspelling correction task as masked word prediction task.

BERT can always be fine-tuned with new training corpus. If the BERT model can be trained on some texts related to the topic of the chosen data set (for instance, some exam scripts for the same questions without errors), there might still be room for further improvement.

Other BERT-like models trained on larger corpus are likely to perform better due to the larger vocabulary size and richer topics of the texts. However, models like Generative Pre-trained Transformer 3 (GPT-3) \cite{b25} require enormous computation power to do inferences, which are less practical in the use case of misspelling correction.

\subsection{Analysis on Part of Speech of the misspellings}

Among all the testing cases, some of the errors seem easy to be identified and corrected by human --- here are 2 examples:

\begin{center}
example 1: ``\emph{Another point which I think was ennoying (correction: annoying) was the concert hall}''
\end{center}
\begin{center}
example 2: ``\emph{There, I saw the exibitions (correction: exhibitions) and admired the building itself}''
\end{center}

Both corrections are only 1 edit distance away from the misspellings. However, in experiment 1, even if we increase the top-N to 500, BERT and the edit distance algorithm still failed to find the correct answer. BERT provided a list of words that matched the Part of Speech (POS) and made sense in the given context. It is possible that if we masked a word with a certain POS, BERT can be confused due to the lack of context. 

In experiment 2, the misspelling ``exibitions'' in example 2 was corrected as ``exhibition'' which was the single form of the expected correction. According to the context, both forms are acceptable in terms of grammar, but it was counted as a failed case in the evaluation. For certain types of POS, we might need to add extra limitations during the evaluation to eliminate the cases like example 2. 

\subsection{BERT can be tolerant of a certain amount of error in the context}

All the testing cases were written by non-native speakers. Many of sentences contained multiple errors. Not only the masked word itself was a misspelling, but there were other types of errors in the sentence as well. BERT showed great robustness in such situations. To further prove this observation, we even tested some informal sentences from social media and BERT still worked properly. Here are some examples:

\begin{center}
example 1: ``\emph{whos gonna tell (masked) my brother that the mf corona is here already and he needs to stay the [...]\footnote{The word is not displayed in the paper due to its offensive nature} out from society for awhile}''
\end{center}
\begin{center}
example 2: ``\emph{seal the whole area let the people be there anyone wants to come in give them a 12hour deadline after that set up a perimeter around the bagh and let them be one (masked) gets out or in till corona is over}''
\end{center}

The second example was actually a paragraph consisted of multiple sentences. The punctuation was missing, and the spellings were chaotic. Even in such situation, BERT still was able to predict the masked word successfully.

\section{Future Work}

\subsection{Misspellings with larger edit distance}

In experiment 2, only similar words with edit distance under 2 were selected as the candidate words due to the balance between accuracy and computation cost. However, in reality, there are spelling errors with a larger edit distance. For example, sometimes people create abbreviations for the long words based on the pronunciations or the spellings. The method used in experiment 2 may not be capable of correcting such errors with the current algorithms. Future work is required to solve this problem. 

\subsection{Other types of errors}

If misspelling correction can be converted to a masked word prediction task, what about correcting other types of errors? For example, some syntax error correction task can also be viewed as a masked prediction task. Some preliminary experiments were also conducted that considered sentences such as:

\begin{center} 
``\emph{The book is going to change the way people thinking.}''
\end{center}

The example above contained a syntax error. Either the verb ``\emph{thinking}'' was in a wrong form, or, less likely, the words "will be" are missing.  

The pre-trained BERT model was very good at masked word prediction as long as the expected output was one word. With this underlying idea, BERT could become a universal model that is capable of correcting different types of errors. The CLC FCE dataset contains different types of error. Some of them are strictly in one-to-one format (the error was one word and the correction was also one word). It would be interesting to compare the results across different types of error.

\subsection{BERT for both misspelling detection and correction}

In the experiments described in this paper, the misspellings were labeled. Therefore BERT did not need to detect the misspelling before fixing it. In real-world application, misspelling detection and correction are always bounded together. It would be more useful if BERT could also be used for misspelling detection.

One approach that allows BERT to detect anomalies could be: scan each of the word in a given sentence and try to find anomaly based on the BERT prediction probability and the edit distance. The metrics need to be carefully designed and tested.

For every single masked word prediction task, BERT would compute the attention scores for every word in the input sentence. Thus the speed of scanning every word in a sentence is similar to the speed of doing one masked work prediction task. 

\section{Conclusion}

This paper explored the possibility of using the pre-trained BERT model for misspelling correction. We utilized the masked word prediction capability of BERT to do misspelling correction. The test result showed the pre-trained BERT model was capable of fixing spelling errors with the help of the edit distance algorithm. 

During our experiments, we did not train or fine-tune the BERT model. Although the accuracy was not close to the ngram model \cite{b10} or the RNN model\cite{b14}, the pre-trained BERT model proved its potential in the misspelling correction task. 

\section*{Acknowledgment}
We would like to thank Baijian Yang for his valuable and helpful feedback during our research progress and Shinhye Yun for her input in the early stages of this work.

\end{document}